%% file: main.tex
\documentclass[letterpaper, 10 pt, conference]{IEEEtran}
\IEEEoverridecommandlockouts
\usepackage{balance}
\usepackage{cite}
\usepackage{amsmath,amssymb,amsfonts,bbm,mathtools,bm}
\usepackage{lipsum}
\usepackage{hyperref}
\usepackage{paralist}
\usepackage{float}
\usepackage[section]{placeins}
\usepackage[ruled, linesnumbered]{algorithm2e}
\usepackage{booktabs}
\SetAlCapNameFnt{\footnotesize}
\SetAlCapFnt{\footnotesize}

\usepackage{graphicx,subcaption}
\usepackage{textcomp}
\usepackage{xcolor}
\usepackage{comment}
\usepackage{multirow}
\let\labelindent\relax
\usepackage{enumitem}
\setlist[enumerate]{leftmargin=3.5mm}
\setlist[itemize]{leftmargin=3.5mm}

\def\BibTeX{{\rm B\kern-.05em{\sc i\kern-.025em b}\kern-.08em
    T\kern-.1667em\lower.7ex\hbox{E}\kern-.125emX}}

\newcommand{\methodShortName}{SurgIRL}
    
\begin{document}

\title{\LARGE \bf \methodShortName{}: Towards Life-Long Learning for Surgical Automation by Incremental Reinforcement Learning}

\author{
Yun-Jie Ho$^1$, Zih-Yun Chiu$^1$, Yuheng Zhi$^1$, and Michael C. Yip$^1$ \IEEEmembership{Member, IEEE}
\thanks{
$^1$Yun-Jie Ho, Zih-Yun Chiu, Yuheng Zhi, and Michael C. Yip are with the Electrical and Computer Engineering Dept., University of California San Diego, La Jolla, CA 92093 USA. {\tt\small \{y8ho, zchiu, yzhi, yip\}@ucsd.edu}}%
}

\maketitle

\begin{abstract}
\input{sections/abstract}
\end{abstract}

\input{sections/introduction}

\input{sections/related_work}

\input{sections/methods}

\input{sections/experiments_results}

\input{sections/conclusion_future}

\newpage

\clearpage
\balance
\bibliographystyle{IEEEtran}
\bibliography{ref}

\end{document}

%% file: sections/abstract.tex
Surgical automation holds immense potential to improve the outcome and accessibility of surgery. Recent studies use reinforcement learning to learn policies that automate different surgical tasks. However, these policies are developed independently and are limited in their reusability when the task changes, making it more time-consuming when robots learn to solve multiple tasks. Inspired by how human surgeons build their expertise, we train surgical automation policies through Surgical Incremental Reinforcement Learning (SurgIRL). SurgIRL aims to (1) acquire new skills by referring to external policies (knowledge) and (2) accumulate and reuse these skills to solve multiple unseen tasks incrementally (incremental learning). Our SurgIRL framework includes three major components. We first define an expandable knowledge set containing heterogeneous policies that can be helpful for surgical tasks. Then, we propose Knowledge Inclusive Attention Network with mAximum Coverage Exploration (KIAN-ACE), which improves learning efficiency by maximizing the coverage of the knowledge set during the exploration process. Finally, we develop incremental learning pipelines based on KIAN-ACE to accumulate and reuse learned knowledge and solve multiple surgical tasks sequentially. Our simulation experiments show that KIAN-ACE efficiently learns to automate ten surgical tasks separately or incrementally. We also evaluate our learned policies on the da Vinci Research Kit (dVRK) and demonstrate successful sim-to-real transfers.

%% file: sections/introduction.tex
\section{Introduction}

Autonomous robot-assisted surgery has recently attracted more attention due to its potential to improve the efficacy and accessibility of surgery. 
Automation techniques can alleviate the challenges of minimally invasive surgery, including the physical and ergonomic challenges of instrumentation, the training required by practitioners to be competent, as well as the observed increase in procedural time.
However, automating surgery has yet to be fully realized due to its high-stakes and complex nature~\cite{ostrander2024current}, thus requiring further research effort. 

Prior studies focus on developing techniques to automate different surgical tasks, such as suturing~\cite{iyer2013single,sen2016automating,pedram2020autonomous,schwaner2021autonomous,hari2024stitch}, blood suction~\cite{richter2021autonomous,huang2021model,ou2024autonomous}, tissue dissection and retraction~\cite{pore2021learning,oh2023framework}, and endoscopic camera control~\cite{ji2018learning,su2021multicamera,moccia2023autonomous}. 
These techniques include optimization, visual servoing, imitation learning, and reinforcement learning (RL). 
Each method has its own advantages, and RL, due to its less dependence on specific objective functions and high-quality expert demonstrations, has become popular in recent years.

Current efforts to automate surgical tasks by RL are performed independently, with each task learned separately from scratch. 
While the trained policies can solve their assigned tasks, they can hardly be reused to help surgical robots learn unseen tasks. 
This is mainly because of the inflexibility of the RL frameworks used: 
Without accumulating knowledge as more tasks are learned, it takes substantially longer time for surgical robots to automate more complex tasks.

\begin{figure}[t!]
    \centering
    \includegraphics[width=1\linewidth]{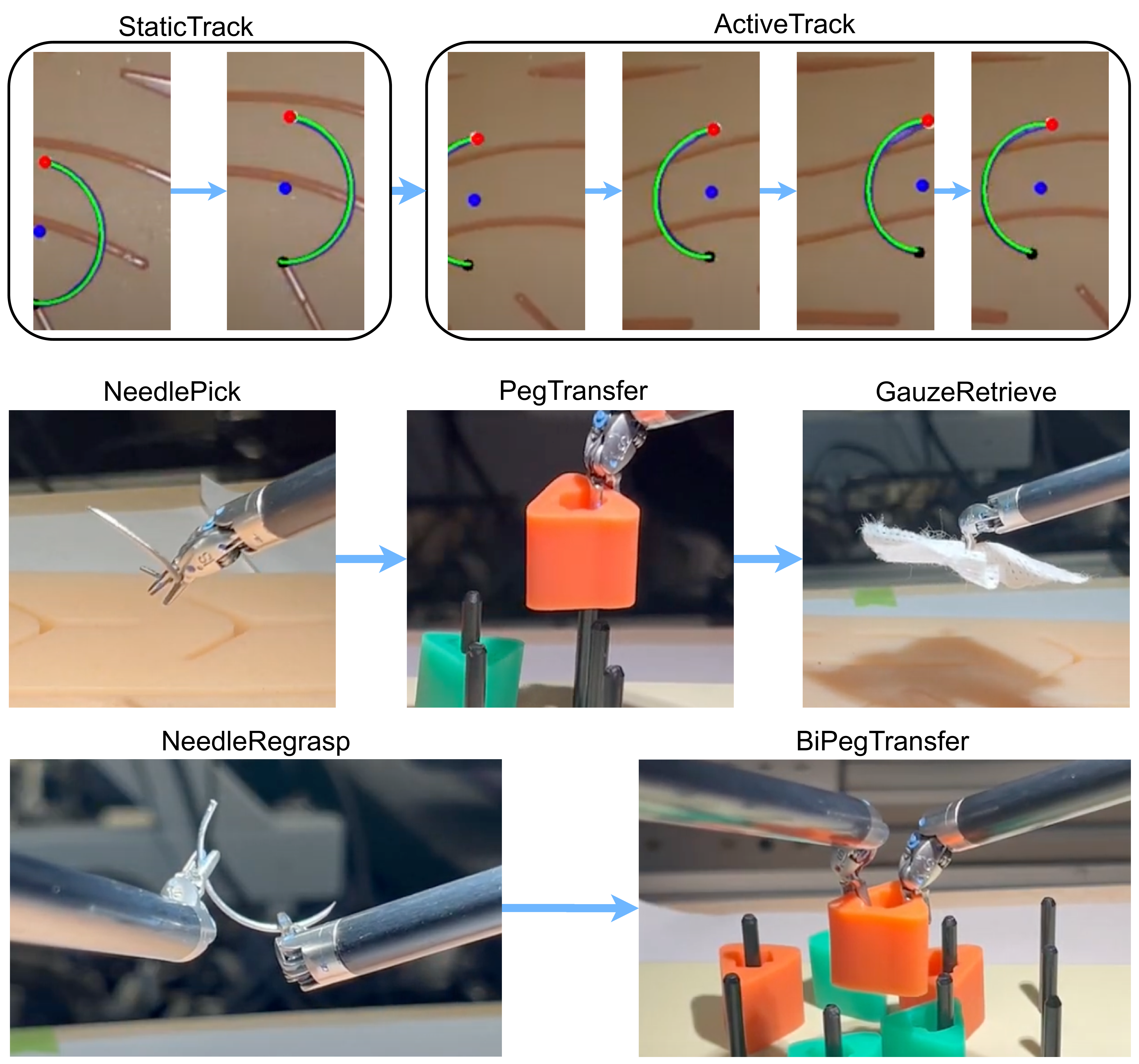}
    \caption{
    The dVRK incrementally learns to automate various surgical tasks, including endoscopic camera control and surgical manipulation. 
    We propose a \methodShortName{} framework that enables surgical robots to learn multiple tasks by accumulating knowledge over tasks with the utmost flexibility. 
    Each row in this figure is a sequence of tasks being incrementally learned. 
    The diversity of the tasks demonstrates the flexibility and effectiveness of our framework.
    }
    \vspace{-5mm}
    \label{fig:cover_figure}
\end{figure}

Human surgeons, however, undergo a more efficient process to gradually develop their expertise. 
They often observe how others perform operations, and these external policies, together with their past experiences, become their knowledge when they perform surgery on their own~\cite{harris2017effect}. 
This learning process is named incremental learning in prior literature~\cite{kaelbling2020foundation,chiu2024flexible} since the knowledge will keep accumulating over time. 
Incremental learning is considered a key to efficient learning due to its flexibility~\cite{kaelbling2020foundation}, i.e., there is no constraint of when and what policies can be used to update the knowledge.

In this work, we explore how a robot can incrementally learn to automate multiple surgical tasks by referring to an expandable knowledge set containing any external policies and previously trained skills.
We propose a Surgical Incremental RL (\methodShortName) framework to achieve this goal. 
\methodShortName\ is a surgical robot learning paradigm that builds upon Knowledge-Grounded RL (KGRL)~\cite{chiu2024flexible}, aiming to achieve surgeon-level learning efficiency and target capabilities such as sample-efficient, generalizable, and incremental learning.
Our \methodShortName\ framework includes three components. 
We first define a knowledge set containing heterogeneous policies that may be helpful for surgical tasks. 
This knowledge set will be incorporated into the learning process and can be arbitrarily expanded with other policies during training.
Next, we propose Knowledge Inclusive Attention Network with mAximum Coverage Exploration (KIAN-ACE) to train surgical agents with guidance from external policies.
KIAN-ACE includes a novel categorical entropy term in the policy-update objective function. 
This term maximizes the usage of the knowledge set when the agents explore environments, leading to better training efficiency.
Finally, we develop surgical incremental learning pipelines based on KIAN-ACE. 
With KIAN-ACE's flexibility to incorporate and accumulate knowledge over tasks, our \methodShortName{} framework can efficiently learn a sequence of surgical tasks that require great precision, high dexterity, and multiple manipulators.

We evaluate our \methodShortName{} agents on ten surgical tasks in simulation and real-world environments. 
The simulation results demonstrate that KIAN-ACE outperforms other methods in learning all the tasks separately. 
Moreover, we show that our incremental learning framework effectively accumulates knowledge to solve multiple unseen tasks. 
It is worth noting that our \methodShortName{} agents can be continuously deployed to learn more surgical tasks. 
Lastly, we run all the trained policies on the real da Vinci Research Kit (dVRK) and show successful sim-to-real transfers (Fig~\ref{fig:cover_figure}). 

Our main contributions are as follows: 
\begin{enumerate}
    \item We propose a \methodShortName\ framework that enables robots to automate multiple surgical tasks incrementally.  
    \item We propose KIAN-ACE as the core \methodShortName\ algorithm and show that it outperforms other methods due to its improved exploration efficiency.  
    \item We apply the learned policies to the real dVRK and show successful sim-to-real transfer on various tasks. 
\end{enumerate}

%% file: sections/related_work.tex
\section{Related Work}

Prior work has developed techniques to automate different surgical procedures.
One line of research focuses on optimizing robot trajectories. 
For example, \cite{sen2016automating,zhong2019dual,pedram2020autonomous,huang2021model,moccia2023autonomous,liang2024real,shinde2024surestep} proposed optimization approaches for surgical tasks such as suturing, needle manipulation, blood suction, endoscope control, and tissue manipulation. 
Another line of research studies how visual information can effectively guide robots to complete those tasks~\cite{iyer2013single,d2018automated,ozguner2021visually,richter2021autonomous,wilcox2022learning,dharmarajan2023automating,oh2023framework,hari2024stitch}.

While optimization and visual servoing methods effectively automate various surgical tasks, they require specific optimization objectives or vision modules. 
To circumvent the need, prior studies have looked into automating surgical tasks by learning from demonstrations (LfD). 
\cite{ji2018learning} leveraged demonstrations to learn features of interest in endoscopic images and guide camera positioning. 
\cite{shin2019autonomous} proposed an LfD-based model predictive control framework to manipulate soft tissues. 
\cite{schwaner2021autonomous} learned surgical action primitives from human demonstrations and composed them to perform suturing. 
\cite{pore2021learning,kim2024surgical,kawaharazuka2024robotic} used imitation learning to learn fundamental tasks such as tissue retraction, knot tying, and peg transfer from expert demonstrations.
LfD enables surgical robots to perform human-like behaviors but requires sufficient high-quality demonstrations, which are difficult to collect.

RL has recently gained popularity due to its less dependence on demonstrations. 
Prior work has proved that RL leads to success in automating surgical tasks such as tissue manipulation~\cite{baek2018path,nguyen2019manipulating,pore2021safe,scheikl2022sim,ou2023sim,shahkoo2023autonomous,karimi2024reward}, needle manipulation~\cite{varier2020collaborative,su2020reinforcement,barnoy2021robotic,chiu2021bimanual,bendikas2023learning,caianiello2023exploring}, camera control~\cite{su2021multicamera}, blood suction~\cite{ou2024autonomous}, and rope cutting~\cite{haiderbhai2024sim2real}. 
In addition, \cite{d2022learning} presented a context-aware RL framework for object pick-and-place in surgical environments. 
\cite{scheikl2021cooperative} showed how multi-agent RL enables surgical robots to cooperate with humans. 
\cite{fan2024learn} studied how to incorporate safety constraints into RL for surgical tasks. 
\cite{huang2023guided,ou2023towards} used expert demonstrations to improve the sample efficiency when training surgical agents with RL. 
Nevertheless, these works focus on using RL to solve different tasks separately without considering how the previously developed skills can be accumulated and reused to build the expertise of surgical agents incrementally. 
Our \methodShortName\ framework takes into account this essential learning behavior, aiming to empower surgical robots with efficient learning over multiple tasks.

%% file: sections/methods.tex
\section{Methods}

Our goal in this work is two-fold: 
\begin{enumerate}
    \item improve autonomous surgical agents' learning efficiency by flexibly incorporating external policies and
    \item accumulate and reuse previously developed policies to learn multiple unseen tasks incrementally. 
\end{enumerate}

We present \methodShortName{}, a paradigm built upon KGRL, to achieve this goal. 
We first discuss the background of KGRL and introduce the major components of our \methodShortName{} framework: an external surgical knowledge set, a policy-learning algorithm that incorporates the knowledge set (KIAN-ACE), and incremental learning pipelines based on KIAN-ACE.

\subsection{Knowledge-Grounded Reinforcement Learning (KGRL)}
\label{subsec:KGRL}
KGRL~\cite{chiu2024flexible} is an RL paradigm studying how agents can efficiently learn by referring to an arbitrary set of external policies. Formally, a KGRL problem is mathematically formulated as a Knowledge-Grounded Markov Decision Process (KGMDP). 
KGMDP is defined by a tuple \( \mathcal{M}_k = (\mathcal{S}, \mathcal{A}, \mathcal{T}, R, \rho, \gamma, \mathcal{G}) \), where \( \mathcal{S} \) is the state space, \( \mathcal{A} \) is the action space, \( \mathcal{T} : \mathcal{S} \times \mathcal{A} \times \mathcal{S} \rightarrow \mathbb{R} \) is the transition probability distribution, \( R \) is the reward function, \( \rho \) is the initial state distribution, \( \gamma \) is the discount factor, and \( \mathcal{G} \) is the set of external knowledge policies. 
KGRL aims to find an optimal policy $\pi^*(\cdot|\cdot;\mathcal{G}) : \mathcal{S} \rightarrow \mathcal{A}$ that maximize the accumulative expected return $\mathbb{E}_{\textbf{s}_0 \sim \rho, \mathcal{T}, \pi^*} \left[\sum_{t=0}^{T} \gamma^t R_t\right]$.

The external knowledge set $\mathcal{G}$ contains $n$ knowledge policies, $\left\{ \pi_{g_1}, \dots, \pi_{g_n} \right\}$. 
A knowledge policy can be any state-action mapping, guiding an agent to explore the environment. 
The external knowledge set should be flexibly expandable and easily shared among different tasks, allowing an agent to leverage the knowledge with the utmost efficiency.

A well-trained KGRL agent is knowledge-acquirable, sample-efficient, generalizable, compositional, and incremental~\cite{chiu2024flexible,kaelbling2020foundation}. 
Our \methodShortName{} framework follows the KGRL formulation and focuses on achieving incremental learning for surgical agents.

\begin{figure}[t!]
    \vspace{1.5mm}
    \centering
    \begin{subfigure}[b]{0.13\textwidth}
    \includegraphics[width=\textwidth]{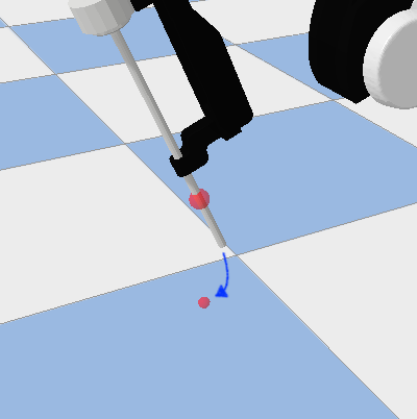}
    \caption{First policy}
    \label{fig:external_kg_demo/kg0}
    \end{subfigure}
    \hspace{8pt}
    \begin{subfigure}[b]{0.13\textwidth}
    \includegraphics[width=\textwidth]{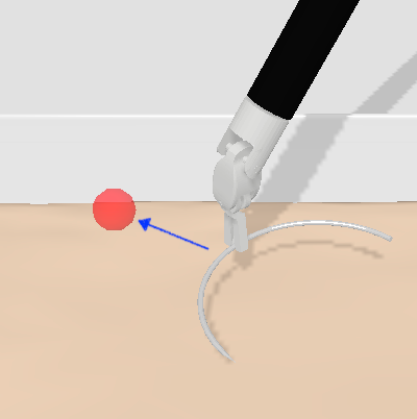}
    \caption{Second policy}
    \label{fig:external_kg_demo/kg1}
    \end{subfigure}
    \hspace{8pt}
    \begin{subfigure}[b]{0.13\textwidth}
    \includegraphics[width=\textwidth]{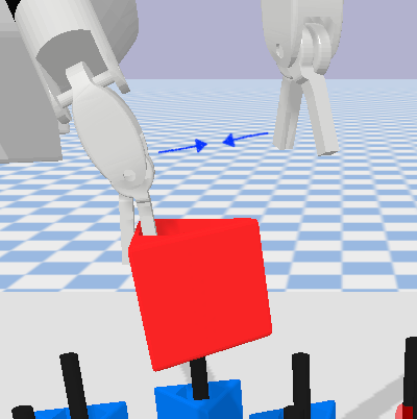}
    \caption{Third policy}
    \label{fig:external_kg_demo/kg2}
    \end{subfigure}
    
    \caption{Visualization of three external knowledge policies for surgical tasks.
    The first policy guides a surgical manipulator to approach an object or move to a point.
    The second policy moves an arm with an object in hand toward a target. 
    The third policy involves two arms trying to hand over an object.
    }
    \label{fig:external_kg_demo}
    \vspace{-3.5mm}
\end{figure}

\subsection{Surgical Knowledge Set}
\label{subsec:surgical-knowledge-set}
\methodShortName{} includes an expandable external knowledge set, $\mathcal{G}$, to guide surgical policy learning. 
We define the initial $\mathcal{G}$ to contain three policies that can be helpful for surgical manipulation.
The first policy guides a surgical manipulator to approach an object or move to a point. 
The second policy moves an arm with an object in hand toward a target. 
The third policy involves two arms approaching each other and trying to hand over an object.  
Fig.~\ref{fig:external_kg_demo} visualizes these policies.

In this work, the initial number of external policies for all tasks is set to three. 
Note that these external policies can be irrelevant to the surgical tasks we considered. 
For example, the second and third external policies do not lead to success in endoscopic camera control tasks since there is no object involved. 
However, irreverent knowledge policies can help an agent explore the environments, and unifying the external knowledge sets makes it easier to share and accumulate knowledge over tasks.

\subsection{Knowledge Inclusive Attention Network with mAximum Coverage Exploration (KIAN-ACE)}
We propose KIAN-ACE as the core learning algorithm of our \methodShortName{} framework. 
KIAN-ACE leverages the external knowledge set and learns a policy that can be incrementally reused to learn unseen tasks. 
KIAN-ACE builds upon Knowledge-Inclusive Attention Network (KIAN)~\cite{chiu2024flexible} and improves its exploration efficiency.

\begin{figure}
    \vspace{1.5mm}
    \centering
    \includegraphics[width=0.475\textwidth]{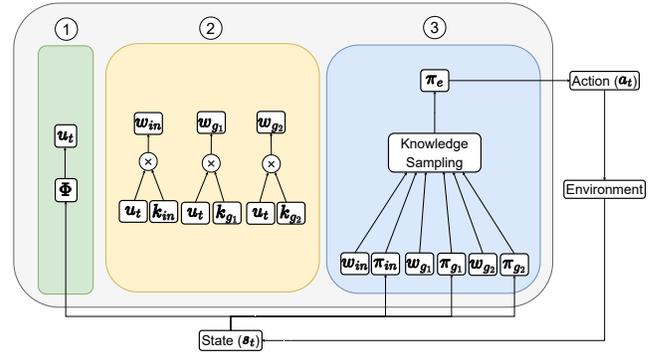}
    \caption{The policy architecture of KIAN-ACE~\cite{chiu2024flexible}. 
    Given a state $\mathbf{s}_t$, the query $\Phi$ outputs a vector $\mathbf{u}_t$. Then, $\mathbf{u}_t$ \textit{attends} each knowledge key, $\mathbf{k}_{in}, \mathbf{k}_{g_1}, \dots$, to calculate the weight of each policy. These weights are used to perform knowledge sampling among all policies, $\pi_{in}, \pi_{g_1}, \dots$. Finally, an action is generated from the sampled policy, $\pi_e$. 
    }
    \label{fig:KIAN_architecture}
    \vspace{-3.5mm}
\end{figure}

KIAN-ACE adopts the same policy architecture as KIAN, as illustrated in Fig.~\ref{fig:KIAN_architecture}. 
A policy has four components: an external knowledge set, an inner actor, knowledge keys, and a query. 
The external knowledge set contains surgical policies introduced in Section~\ref{subsec:surgical-knowledge-set} and is expandable with other policies. 
The inner actor, $\pi_{in}(\cdot|\cdot;\theta):\mathcal{S}\rightarrow\mathcal{A}$, is a learnable function approximator that enables an agent to develop skills different from the external ones. 
Each external or internal policy is paired with a state-independent knowledge key, $\textbf{k}_{in}$ or $\textbf{k}_{g_j}$, which is a $d_k$-dimensional learnable embedding that represents the policy. 
Knowledge keys encode all policies in a unified representation space, allowing the knowledge set to consist of policies in diverse forms. 
Finally, the query, $\Phi(\cdot; \phi):\mathcal{S} \rightarrow \mathbb{R}^{d_k}$, is a learnable function approximator that outputs a $d_k$-dimensional vector, $\mathbf{u}_t$, given a state. 
This output and the knowledge keys will be used to perform \textit{knowledge sampling} during training.

KIAN-ACE generates an action through an embedding-based attention mechanism~\cite{chiu2024flexible}. 
This attention mechanism predicts the weight of each policy as follows:
{\small
\begin{align}
    \mathbf{u}_t = \Phi(\mathbf{s}_t; \phi),\ w_{t,in} = (\mathbf{u}_t \cdot \mathbf{k}_{in})/c_{t,in}, &\\
    w_{t,g_j} = (\mathbf{u}_t \cdot \mathbf{k}_{g_j})/c_{t,g_j}, \ \forall j \in \{1, \dots, n\}, &
\end{align}
}%
where $c_{t,in} \in \mathbb{R}$ and $c_{t,g_j} \in \mathbb{R}$ are normalization factors.
Then, an agent performs \textit{knowledge sampling} to select a policy based on these weights using Gumbel softmax~\cite{jang2016categorical}:
{\small
\begin{equation}
    e \sim \texttt{gumbel\_softmax}([w_{t,in}, w_{t,g_1}, \dots, w_{t,g_n}]^\top).
\end{equation}
}%
Finally, an action is sampled from $\pi_e(\cdot | \mathbf{s}_t)$. 
Note that all learnable components, including the inner actor, knowledge keys, and the query, can be updated by any policy-gradient algorithm. 
This work uses maximum entropy algorithms such as Soft Actor-Critic (SAC)~\cite{haarnoja2018soft} to encourage exploration.


The major distinction between KIAN-ACE and KIAN is their policy-update objectives, which lead to different exploration strategies.
The objective of KIAN is 
{\small
\begin{align}
    & \pi^* = \arg\max_{\pi} \sum_{t} \mathbb{E}_\pi \left[ R_t
            + \alpha H(\pi(\cdot | \mathbf{s}_t)) \right], \\
    & H(\pi(\cdot | \mathbf{s}_t)) \approx \sum_i \pi(\mathbf{a}_i | \mathbf{s}_t) \log \pi(\mathbf{a}_i | \mathbf{s}_t), 
\end{align}
}%
where $\alpha \in \mathbb{R}$ is a hyperparameter, and $H(\pi(\cdot | \mathbf{s}_t))$ is the entropy of the policy approximated by $i$ action samples. 
In maximum-entropy exploration, KIAN purely relies on maximizing the randomness of the sampled actions, which can lead to the weights, $\left[w_{t,in}\ w_{t,g_1} \dots\ w_{t,g_n}\right]$, being highly biased.
This is more likely to happen if the knowledge policies have very different entropy, which is termed \textit{entropy imbalance} in~\cite{chiu2024flexible}.
Biased weights result in an agent not leveraging the knowledge set enough to achieve efficient learning. 
Therefore, we propose also to maximize \textit{the randomness of knowledge sampling} during maximum-entropy exploration.
The objective of KIAN-ACE becomes
{\small
\begin{align}
    & \pi^* = \arg\max_{\pi} \sum_{t} \mathbb{E}_\pi \left[ R_t
            + \alpha H(\pi(\cdot | \mathbf{s}_t)) + \beta H(\mathbf{w}_t) \right], \label{eq:KIAN-ACE-objective}\\
    & H(\mathbf{w}_t) = -w_{t,in} \cdot \log w_{t,in} -\sum_{j=1}^n w_{t,g_j} \cdot \log w_{t,g_j}, 
\end{align}
}%
where $\beta \in \mathbb{R}$ is a hyperparameter. 
By adding the term $H(\mathbf{w}_t)$, the weights of all knowledge policies become more uniform.
Uniform weights ensure that an agent \textit{maximizes the usage (coverage) of the knowledge set} during exploration.

The value of $\beta$ in equation (\ref{eq:KIAN-ACE-objective}) determines how much an agent will explore the environment. 
Higher $\beta$ favors exploration, and lower $\beta$ favors exploitation. 
As an agent starting to master the task, $\beta$ should decrease to avoid unnecessary exploration.
We propose to decay $\beta$ exponentially as follows: 
{\small
\begin{equation}
    \beta_t = \exp(-d_e \times t) + c_e, 
    \label{eq:beta}
\end{equation}
}%
where $d_e \in \mathbb{R}$ is the decay rate, and $c_e \in \mathbb{R}$ is a constant deciding the base coefficient of $H(\hat{\mathbf{w}}_t)$. 
Equation (\ref{eq:beta}) leads to better training stability since an agent can adjust the level of exploration by changing $d_e$.
For instance, higher $d_e$ better suits simple environments as they require less exploration to learn good strategies.
Lastly, non-zero $c_e$ ensures some randomness throughout the training process, leaving room for further improved performance.

\subsection{Incremental Learning for Surgical Tasks}
\label{subsec:incremental-learning-pipelines}
In our \methodShortName{} framework, we present incremental learning pipelines based on KIAN-ACE to accumulate knowledge and efficiently learn multiple surgical tasks. 
The essential behavior of knowledge accumulation is that an agent can reuse (1) the previously developed policies and (2) \textit{how the external knowledge set is leveraged}.

In KIAN-ACE, the knowledge that can be accumulated over tasks includes the external/inner policies, the knowledge keys, and the query. 
External and internal policies are previously developed skills that can help explore new environments. 
A knowledge key encodes a policy to a representation space shared by all policies, enabling an agent to identify their similarities. 
A query informs an agent which policies are more helpful for a task. 
Knowledge keys and queries together guide an agent to navigate a knowledge set.

\begin{figure}[t!]
    \vspace{1.5mm}
    \centering
    \begin{subfigure}[b]{\linewidth}
    \centering
    \includegraphics[width=\linewidth]{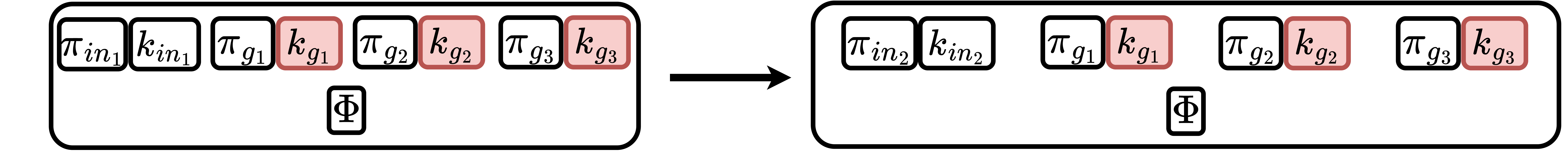}
    \caption{Reusing the knowledge keys}
    \label{fig:reuse_keys}
    \end{subfigure}

    \vspace{10pt}

    \begin{subfigure}[b]{\linewidth}
    \centering
    \includegraphics[width=\linewidth]{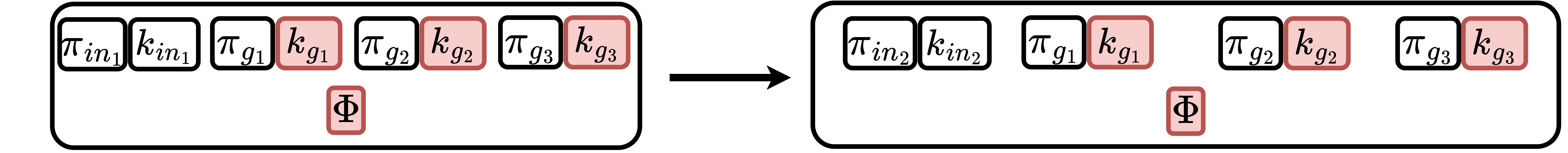}
    \caption{Reusing the knowledge keys and query}
    \label{fig:reuse_query_keys}
    \end{subfigure}

    \vspace{10pt}

    \begin{subfigure}[b]{\linewidth}
    \centering
    \includegraphics[width=\linewidth]{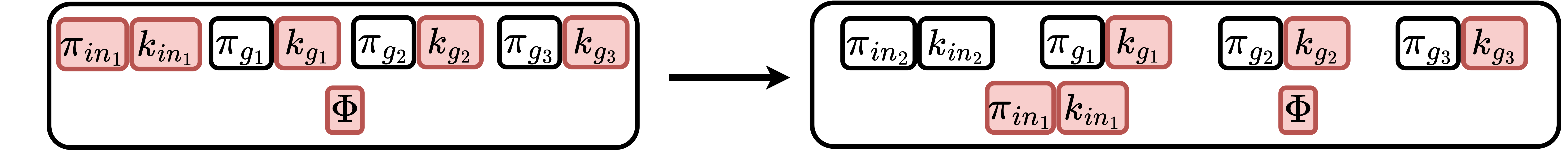}
    \caption{Reusing the knowledge keys, query, and internal policy}
    \label{fig:reuse_all}
    \end{subfigure}

    \caption{
    \methodShortName{}'s three incremental learning pipelines.
    The red blocks indicate the components reused from one task to another. 
    Each pipeline has its best-fitted scenarios based on state/action spaces and similarity of tasks. 
    Fig.~\ref{fig:reuse_keys} suits tasks with different state/action spaces. 
    Fig.~\ref{fig:reuse_query_keys} suits tasks with the same state/action spaces but with different environmental dynamics. 
    Fig.~\ref{fig:reuse_all} suits tasks with greater similarity.
    }
    \label{fig:incremental_setup}
    \vspace{-3.5mm}
\end{figure}

We propose three incremental learning pipelines that reuse different components of KIAN-ACE to achieve knowledge accumulation. 
Each pipeline has its best-suited scenarios based on the state/action space and similarity of tasks.
\begin{enumerate}
    \item For tasks with different observation/action spaces, an agent can \textbf{only reuse knowledge keys}, which are state-independent, to acquire the relationships between policies.
    \item For tasks with the same space but with different environmental dynamics, an agent can \textbf{reuse knowledge keys and queries} to efficiently navigate a knowledge set.
    \item For tasks with greater similarity, an agent can \textbf{reuse all components} to take full advantage of prior knowledge.
\end{enumerate}
Fig.~\ref{fig:incremental_setup} visualizes the increment learning pipelines. 
Note that there is no limitation on the number of tasks learned incrementally or their order. 
The freedom to learn any sequence of tasks highlights the flexibility of our \methodShortName{} framework.


%% file: sections/experiments_results.tex
\section{Experiments and Results} 

We evaluate our \methodShortName{} framework on ten surgical tasks~\cite{xu2021surrol}. 
These tasks include (1) controlling surgical manipulators, such as a patient side manipulator (PSM) and an endoscopic camera manipulator (ECM), (2) object manipulation, such as gauze retrieval, needle pickup, peg transfer, bimanual needle regrasping, and bimanual peg transfer, and (3) camera viewpoint control and object tracking.

\subsection{Simulation Experiments}
\begin{figure*}[t!]
  \vspace{1.5mm}
  \centering
    \includegraphics[width=1\textwidth]{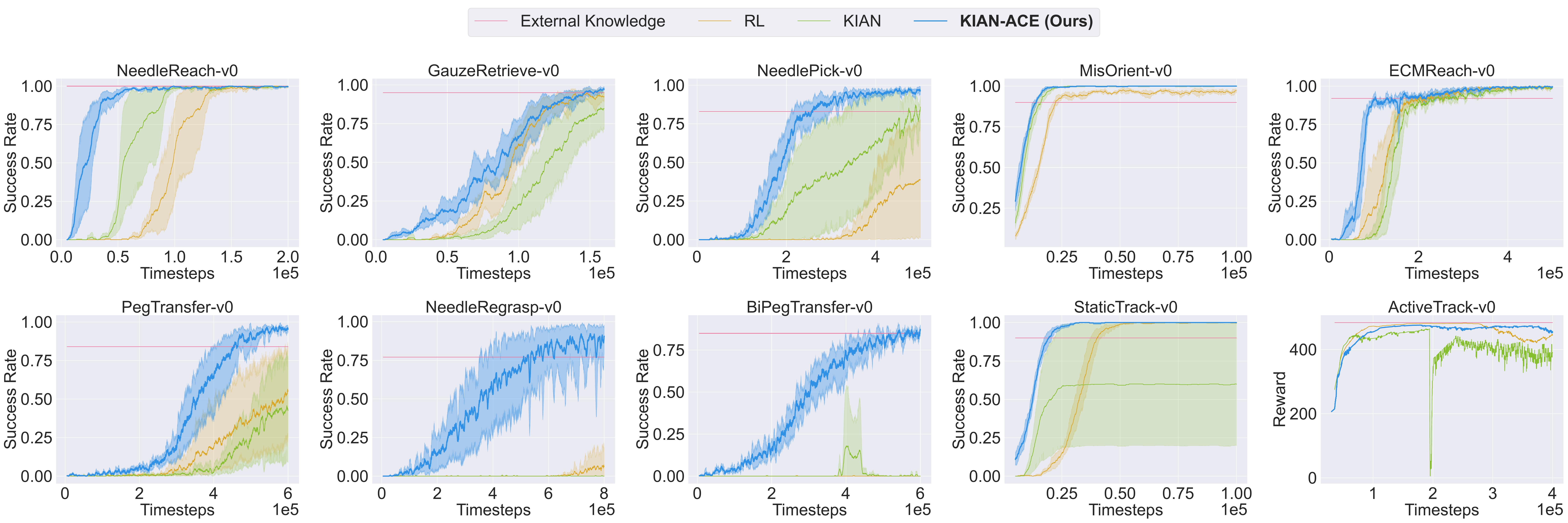}
  \caption{The performance of external policies, RL~\cite{haarnoja2018soft}, KIAN~\cite{chiu2024flexible}, and KIAN-ACE in single-task learning. KIAN-ACE outperforms other methods and has more consistent training results, demonstrating the effectiveness of its exploration strategy.}
  \label{fig:comparisons_of_all}
  \vspace{-3.5mm}
\end{figure*}

\begin{table*}[t!]
\centering
\caption{Incremental Learning Experimental Setup}
\begin{tabular}{cccc}
\toprule
\textbf{Group} & \textbf{Task Sequence} & \textbf{Task Similarity} & \textbf{Incremental Learning Pipeline} \\ 
\midrule
1 & MisOrient, ECMReach, NeedleReach & Different observation/action spaces & Reuse knowledge keys (Fig.~\ref{fig:reuse_keys}) \\
\midrule
\multirow{2}{*}{2} & \multirow{2}{*}{NeedleRegrasp, BiPegTransfer} & Same observation/action spaces & \multirow{2}{*}{ Reuse knowledge keys and query (Fig.~\ref{fig:reuse_query_keys})} \\
 & & but different environmental dynamics & \\
\midrule
\multirow{2}{*}{3} & NeedlePick, PegTransfer & \multirow{2}{*}{Greater Similarity} & Reuse all components (Fig.~\ref{fig:reuse_all}) \\
 & GauzeRetrieve & & GauzeRetrieve does not reuse inner actors \\
\midrule
4 & StaticTrack, ActiveTrack & Greater Similarity & Reuse all components (Fig.~\ref{fig:reuse_all}) \\
\bottomrule
\end{tabular}
\label{tab:incremental-learning-setup}
\vspace{-5mm}
\end{table*}

\subsubsection{Experimental setup}
We train \methodShortName{} agents in SurRoL~\cite{xu2021surrol}, a simulation platform for surgical robot learning. 
We modify some environments in SurRoL for simulation and training stability. 
In GauzeRetrieve, NeedlePick, PegTransfer, NeedleRegrasp, and BiPegTransfer, a policy does not control whether a gripper is closed/opened. 
A gripper is automatically closed whenever its distance to an object is smaller than a threshold, which ranges from 0.01 to 0.1 in different tasks. 
Similarly, a gripper is automatically opened whenever the distance between an object and the goal is smaller than 0.01, or the other gripper holds the object.
Additionally, we design a dense reward function as follows: 

\vspace{-5mm}
{\small
\begin{equation}
    R_t = -c_{og} \cdot d_{t,og} - c_{ro} \cdot d_{t,ro} - c_{rg} \cdot d_{t,rg} - p + 20 \cdot success, 
\end{equation}
}
where $d_{t,og}, d_{t,ro}$, and $d_{t,rg} \in \mathbb{R}$ are the object-goal distance, robot-object distance, and robot-goal distance, respectively, $c_{og}, c_{ro}$, and $c_{rg} \in \mathbb{R}$ are the coefficients of these distances, $p \in \mathbb{R}_{\geq 0}$ is the penalty when collision happens, and $success$ indicates whether the task is successfully completed. 
The values of $c_{og}, c_{ro}$, and $c_{rg}$ lies within $\{0, 1\}$, and $p \in \{0, 2\}$.

We compare the performance of four different algorithms: external policies as introduced in Section~\ref{subsec:surgical-knowledge-set}, RL (SAC~\cite{haarnoja2018soft}), KIAN~\cite{chiu2024flexible}, and our KIAN-ACE.
KIAN and KIAN-ACE use SAC to perform actor-critic policy learning.
The actor architecture of RL and the inner actor architectures of KIAN and KIAN-ACE are multi-layer perceptrons (MLPs) with three hidden layers and a hidden size of 512 units. 
The dimension of each knowledge key $d_k = 4$. 
The query network is an MLP with three hidden layers and a hidden size of 64 units.
The actor learning rate ranges from $5 \times 10^{-5}$ to $3 \times 10^{-4}$. 
The entropy coefficient $\alpha$ lies within $\{10^{-1}, 10^{-2}, 10^{-3}, 10^{-5}\}$ or can be automatically tuned~\cite{haarnoja2018soft}, and $\beta$ lies within $\{0, 2\times10^{-4}\}$.

\subsubsection{Single-Task Learning}
Fig.~\ref{fig:comparisons_of_all} shows the learning curves of each task trained separately. 
Each error band in Fig.~\ref{fig:comparisons_of_all} represents the $95\%$ confidence interval. 
KIAN-ACE achieves the best performance and sample efficiency in all tasks. 
The improvement is more evident in challenging tasks such as NeedleRegrasp and BiPegTransfer. 
In addition, the error bands of KIAN-ACE are smaller than those of other methods, indicating that its training results are more consistent.

\subsubsection{Incremental Learning}
We evaluate \methodShortName{}'s incremental learning pipelines, as introduced in Section~\ref{subsec:incremental-learning-pipelines}, by dividing the ten surgical tasks into four groups. 
Table~\ref{tab:incremental-learning-setup} shows each group's task sequence, task similarity, and the incremental learning pipeline used. 
The policy of the first task in each group comes from Fig.~\ref{fig:comparisons_of_all}.
We perform the following modifications to the incremental learning pipelines or environments based on our understanding of the surgical tasks: 
For tasks in Group 2, we only reuse $\mathbf{k}_{g_1}$ and $\mathbf{k}_{g_2}$ instead of all external knowledge keys due to the last external policies of these tasks being inconsistent. 
For GauzeRetrieve in Group 3, we do not reuse the inner actors of previous tasks since the object dynamics of a gauze is very different from that of a needle or a cube. 
For tasks in Group 4, we unify the state space of StaticTrack and ActiveTrack so that the query and inner actor can be reused between them.

\begin{figure}[t!]
    \centering
    \includegraphics[width=0.475\textwidth]{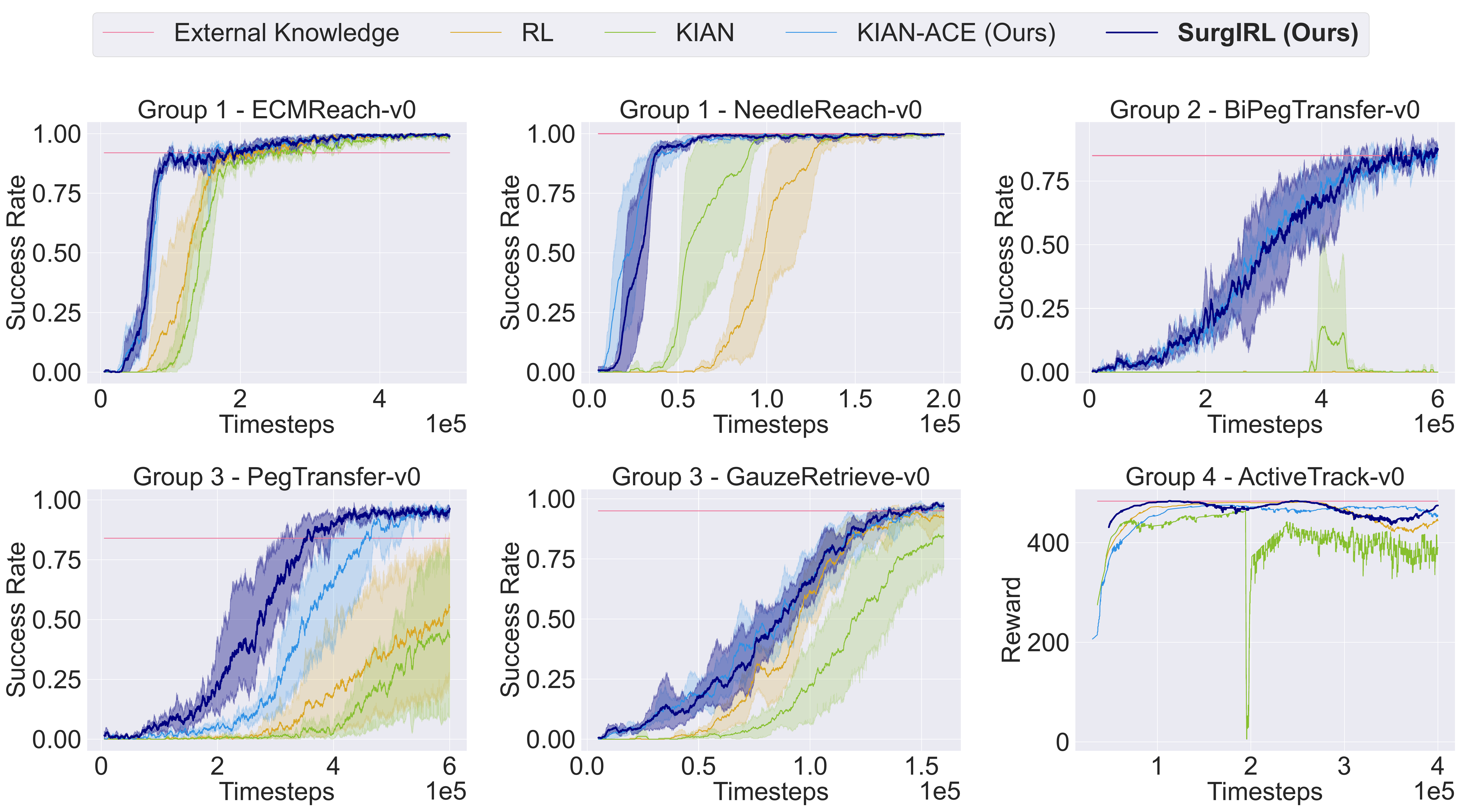}
    \caption{
    Incremental learning results. Overall, \methodShortName{} performs better or comparable. The improvement is more evident for similar tasks. Note that our \methodShortName{} agents can be continuously improved by training in more surgical tasks. 
    }
    \label{fig:IL_all}
    \vspace{-6.5mm}
\end{figure}

Fig.~\ref{fig:IL_all} shows the learning curves of each task trained incrementally. 
Our \methodShortName{} framework successfully learns multiple surgical tasks in sequence and achieves better or similar results compared to other single-task agents. 
For tasks with greater similarity, such as NeedlePick and PegTransfer in Group 3 and StaticTrack and ActiveTrack in Group 4, incremental learning can effectively improve the training efficiency. 
Otherwise, an agent needs enough interaction samples to succeed in an unseen task. 
Moreover, if the environmental dynamics in two tasks differ, such as NeedleRegrasp and BiPegTransfer in Group 2 and PegTransfer and GauzeRetrieve in Group 3, reusing the query does not always provide helpful information on navigating the knowledge set, leading to comparable results. 
However, our \methodShortName{} agents can be continuously improved by training in more surgical tasks, opening the door to life-long surgical learning.

\subsection{Real-Robot Experiments}

We deploy the trained \methodShortName{} policies on a real dVRK~\cite{kazanzides2014open} to demonstrate the sim-to-real transferability. 
A PSM attaches a Large Needle Driver (LND) that interacts with the objects (suture needles, gauze, or blocks). 
An ECM holds a stereo endoscope, which is 1080p and runs at 30 fps.

We run prior visual tracking methods to provide an input state, $\mathbf{s}_t$, to a policy. 
The PSM's end-effectors and needles are tracked with~\cite{richter2021robotic,chiu2022markerless,chiu2023real}. 
We use Cutie~\cite{cheng2024putting} to obtain real-time needle detections, with an initial segmentation extracted from the Segment Anything Model (SAM)~\cite{kirillov2023segment}. 
The tracking algorithms run at 20 fps. 
The gauze and blocks are detected using ArUco markers~\cite{garrido2014automatic}. 
The ECM's end-effector is detected in its base frame using forward kinematics.

The poses of the PSM's end-effectors and objects are detected in the \textit{camera} frame, but the policies trained in SurRoL take input in a \textit{robot's base} frame. 
Therefore, we use the following equation to transform a pose in the camera frame to a pose in a robot's base frame: 
{\footnotesize
\begin{align}
    \mathbf{H}^b_e\left( \mathbf{p}^b_e, \mathbf{q}^b_e \right) & = \mathbf{H}^c_b\left( \mathbf{p}^c_b, \mathbf{q}^c_b \right)^{-1} \mathbf{H}^c_e\left( \mathbf{p}^c_e, \mathbf{q}^c_e \right) \\
    \mathbf{H}^b_o\left( \mathbf{p}^b_o, \mathbf{q}^b_o \right) & = \mathbf{H}^c_b\left( \mathbf{p}^c_b, \mathbf{q}^c_b \right)^{-1} \mathbf{H}^c_o\left( \mathbf{p}^c_o, \mathbf{q}^c_o \right),
\end{align}
}
where $\textbf{p}^b_{\{e,o\}} \in \mathbb{R}^3$ and $\textbf{q}^b_{\{e,o\}} \in \mathbb{R}^3$ are the position and orientation of an end-effector or an object in a robot's base frame, $\textbf{p}^c_{\{e,o\}} \in \mathbb{R}^3$ and $\textbf{q}^c_{\{e,o\}} \in \mathbb{R}^3$ are the position and orientation of an end-effector or an object in the camera frame, $\textbf{p}^c_b \in \mathbb{R}^3$ and $\textbf{q}^c_b \in \mathbb{R}^3$ are the position and orientation of a robot base in the camera frame, and $\mathbf{H}\left( \mathbf{p}, \mathbf{q} \right) \in SE(3)$ is the homogeneous transformation matrix of $( \mathbf{p}, \mathbf{q} )$.

\begin{table}[t]
\centering
\caption{Performance of \methodShortName{} agents on the real dVRK} 
\label{tab:real-robot-experiments}
\begin{tabular}{ccccccc}
\toprule
     & \multicolumn{2}{c}{GauzeRetrieve} & \multicolumn{2}{c}{NeedlePick} & \multicolumn{2}{c}{PegTransfer} \\
    Success Rate & \multicolumn{2}{c}{20 / 20} & \multicolumn{2}{c}{19 / 20} & \multicolumn{2}{c}{20 / 20} \\
    \midrule
     & \multicolumn{3}{c}{NeedleRegrasp} & \multicolumn{3}{c}{BiPegTransfer} \\
    Success Rate & \multicolumn{3}{c}{18 / 20} & \multicolumn{3}{c}{20 / 20} \\
    \midrule
     & \multicolumn{3}{c}{StaticTrack} & \multicolumn{3}{c}{ActiveTrack} \\
    Position Distance (mm) & \multicolumn{3}{c}{$2.22 \pm 0.54$} & \multicolumn{3}{c}{$3.22 \pm 0.34$} \\
    Pixel Distance & \multicolumn{3}{c}{$29.86 \pm 12.91$} & \multicolumn{3}{c}{$28.31 \pm 13.82$} \\
\bottomrule
    \end{tabular}
 \vspace{-6mm}
\end{table}

\begin{figure}[t!]
    \vspace{1.5mm}
    \centering
    \includegraphics[width=0.475\textwidth]{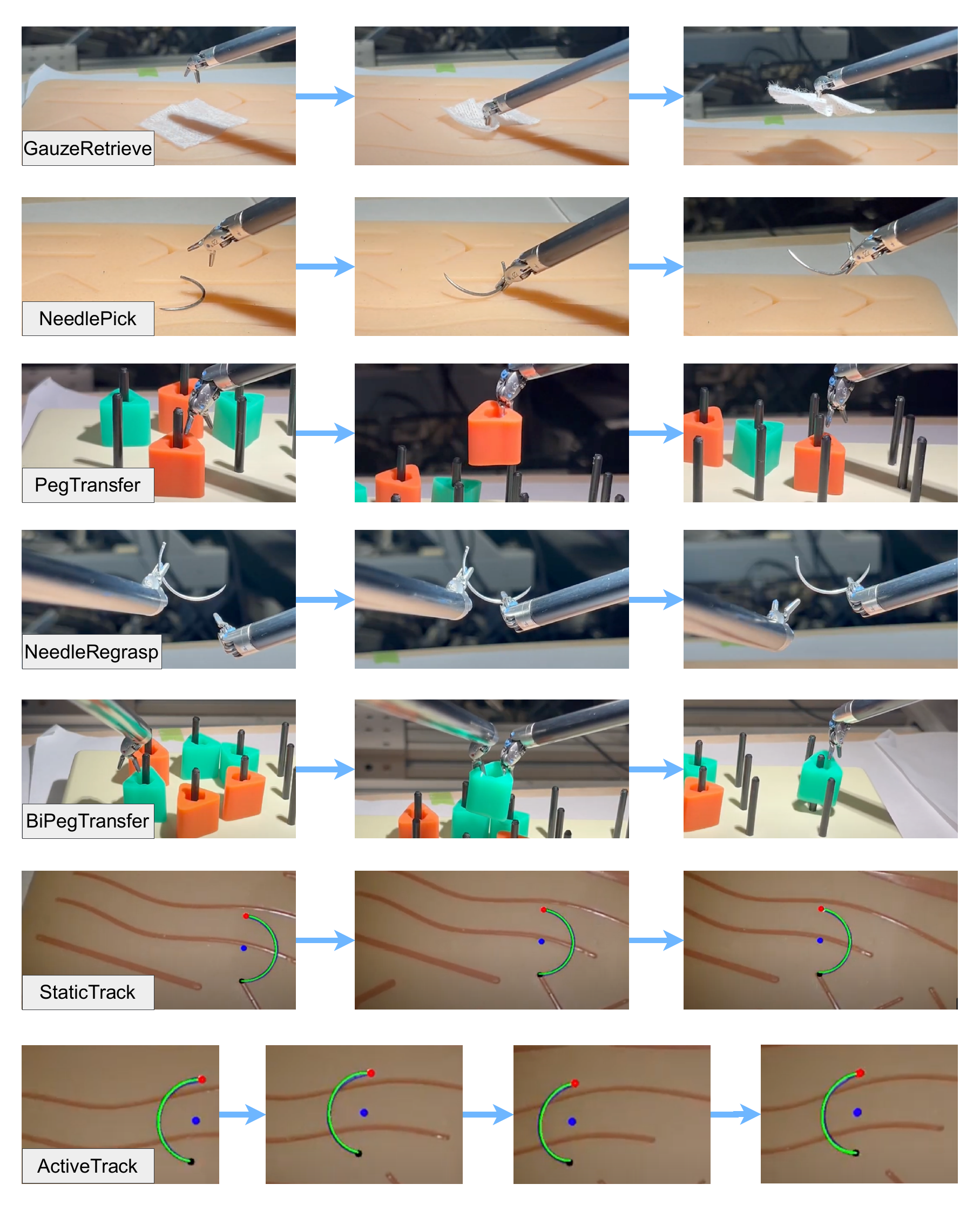}
    \caption{Deployment of our \methodShortName{} agents in Fig.~\ref{fig:IL_all} on the real dVRK. We demonstrate successful sim-to-real transfers of our \methodShortName{} agents across diverse surgical tasks.}
    \label{fig:real_robot_experiment}
    \vspace{-4.5mm}
\end{figure}

We evaluate the best-performed models in Fig.~\ref{fig:IL_all} on the real dVRK, and Table~\ref{tab:real-robot-experiments} shows their success rate or distance to the goal. 
We run 20 trials for each task. 
Note that we do not specifically evaluate MisOrient, ECMReach, and NeedleReach since they are subtasks of other tasks. 
Our \methodShortName{} policies achieve over 90\% success rate on all PSM tasks, demonstrating effective sim-to-real transfers. 
For ECM tasks, the goal is to move the camera such that a suture needle appears at the center of an image. 
After the ECM's motions are stabilized, we record the camera's distance to the goal in the 3D and image spaces. 
The positional errors are below 4mm for all trails, and the pixel errors are below 40 for StaticTrack and below 60 for ActiveTrack. 
The main reasons for these errors are (1) the goal might be out of the ECM's reachable space and (2) the ECM's motors do not provide enough torque to reach the desired pose. 
However, these errors are not significant, and overall, the camera properly centers the object. 
Fig.~\ref{fig:cover_figure} and \ref{fig:real_robot_experiment} demonstrate that our \methodShortName{} agents succeed in automating multiple surgical tasks.

%% file: sections/conclusion_future.tex
\section{Conclusion and Future work}

We introduce \methodShortName{}, a surgical incremental RL paradigm aiming to (1) learn new surgical tasks by referring to an arbitrary set of external policies and (2) accumulate and reuse these skills to solve multiple unseen tasks. 
We propose a \methodShortName{} framework containing external surgical policies, a learning algorithm, KIAN-ACE, leveraging external policies to improve exploration efficiency, and incremental learning pipelines based on KIAN-ACE to accumulate knowledge and learn any sequence of surgical tasks.
We demonstrate the effectiveness of our \methodShortName{} agents in simulation and successfully transfer them to the real dVRK.

However, our \methodShortName{} framework still leaves room for improvement. 
Future work includes enhancing the incremental learning algorithm to better (1) handle tasks with different observation/action spaces and environmental dynamics and (2) filter out non-helpful knowledge policies while retraining informative ones. 
This work serves as a first step in surgical incremental learning, and we hope it opens the door toward building life-long surgical agents.

%% file: main.bbl
\begin{thebibliography}{10}
\providecommand{\url}[1]{#1}
\csname url@samestyle\endcsname
\providecommand{\newblock}{\relax}
\providecommand{\bibinfo}[2]{#2}
\providecommand{\BIBentrySTDinterwordspacing}{\spaceskip=0pt\relax}
\providecommand{\BIBentryALTinterwordstretchfactor}{4}
\providecommand{\BIBentryALTinterwordspacing}{\spaceskip=\fontdimen2\font plus
\BIBentryALTinterwordstretchfactor\fontdimen3\font minus \fontdimen4\font\relax}
\providecommand{\BIBforeignlanguage}[2]{{%
\expandafter\ifx\csname l@#1\endcsname\relax
\typeout{** WARNING: IEEEtran.bst: No hyphenation pattern has been}%
\typeout{** loaded for the language `#1'. Using the pattern for}%
\typeout{** the default language instead.}%
\else
\language=\csname l@#1\endcsname
\fi
#2}}
\providecommand{\BIBdecl}{\relax}
\BIBdecl

\bibitem{ostrander2024current}
B.~T. Ostrander, D.~Massillon, L.~Meller, Z.-Y. Chiu, M.~Yip, and R.~K. Orosco, ``The current state of autonomous suturing: a systematic review,'' \emph{Surgical Endoscopy}, vol.~38, no.~5, pp. 2383--2397, 2024.

\bibitem{iyer2013single}
S.~Iyer, T.~Looi, and J.~Drake, ``A single arm, single camera system for automated suturing,'' in \emph{2013 IEEE international conference on robotics and automation}.\hskip 1em plus 0.5em minus 0.4em\relax IEEE, 2013, pp. 239--244.

\bibitem{sen2016automating}
S.~Sen, A.~Garg, D.~V. Gealy, S.~McKinley, Y.~Jen, and K.~Goldberg, ``Automating multi-throw multilateral surgical suturing with a mechanical needle guide and sequential convex optimization,'' in \emph{2016 IEEE international conference on robotics and automation (ICRA)}.\hskip 1em plus 0.5em minus 0.4em\relax IEEE, 2016, pp. 4178--4185.

\bibitem{pedram2020autonomous}
S.~A. Pedram, C.~Shin, P.~W. Ferguson, J.~Ma, E.~P. Dutson, and J.~Rosen, ``Autonomous suturing framework and quantification using a cable-driven surgical robot,'' \emph{IEEE Transactions on Robotics}, vol.~37, no.~2, pp. 404--417, 2020.

\bibitem{schwaner2021autonomous}
K.~L. Schwaner, I.~Iturrate, J.~K. Andersen, P.~T. Jensen, and T.~R. Savarimuthu, ``Autonomous bi-manual surgical suturing based on skills learned from demonstration,'' in \emph{2021 IEEE/RSJ International Conference on Intelligent Robots and Systems (IROS)}.\hskip 1em plus 0.5em minus 0.4em\relax IEEE, 2021, pp. 4017--4024.

\bibitem{hari2024stitch}
K.~Hari, H.~Kim, W.~Panitch, K.~Srinivas, V.~Schorp, K.~Dharmarajan, S.~Ganti, T.~Sadjadpour, and K.~Goldberg, ``Stitch: Augmented dexterity for suture throws including thread coordination and handoffs,'' \emph{arXiv preprint arXiv:2404.05151}, 2024.

\bibitem{richter2021autonomous}
F.~Richter, S.~Shen, F.~Liu, J.~Huang, E.~K. Funk, R.~K. Orosco, and M.~C. Yip, ``Autonomous robotic suction to clear the surgical field for hemostasis using image-based blood flow detection,'' \emph{IEEE Robotics and Automation Letters}, vol.~6, no.~2, pp. 1383--1390, 2021.

\bibitem{huang2021model}
J.~Huang, F.~Liu, F.~Richter, and M.~C. Yip, ``Model-predictive control of blood suction for surgical hemostasis using differentiable fluid simulations,'' in \emph{2021 IEEE International Conference on Robotics and Automation (ICRA)}.\hskip 1em plus 0.5em minus 0.4em\relax IEEE, 2021, pp. 12\,380--12\,386.

\bibitem{ou2024autonomous}
Y.~Ou, A.~Soleymani, X.~Li, and M.~Tavakoli, ``Autonomous blood suction for robot-assisted surgery: A sim-to-real reinforcement learning approach,'' \emph{IEEE Robotics and Automation Letters}, 2024.

\bibitem{pore2021learning}
A.~Pore, E.~Tagliabue, M.~Piccinelli, D.~Dall’Alba, A.~Casals, and P.~Fiorini, ``Learning from demonstrations for autonomous soft-tissue retraction,'' in \emph{2021 international symposium on medical robotics (ISMR)}.\hskip 1em plus 0.5em minus 0.4em\relax IEEE, 2021, pp. 1--7.

\bibitem{oh2023framework}
K.-H. Oh, L.~Borgioli, M.~Zefran, L.~Chen, and P.~C. Giulianotti, ``A framework for automated dissection along tissue boundary,'' \emph{arXiv preprint arXiv:2310.09669}, 2023.

\bibitem{ji2018learning}
J.~J. Ji, S.~Krishnan, V.~Patel, D.~Fer, and K.~Goldberg, ``Learning 2d surgical camera motion from demonstrations,'' in \emph{2018 IEEE 14th International Conference on Automation Science and Engineering (CASE)}.\hskip 1em plus 0.5em minus 0.4em\relax IEEE, 2018, pp. 35--42.

\bibitem{su2021multicamera}
Y.-H. Su, K.~Huang, and B.~Hannaford, ``Multicamera 3d viewpoint adjustment for robotic surgery via deep reinforcement learning,'' \emph{Journal of Medical Robotics Research}, vol.~6, no. 01n02, p. 2140003, 2021.

\bibitem{moccia2023autonomous}
R.~Moccia and F.~Ficuciello, ``Autonomous endoscope control algorithm with visibility and joint limits avoidance constraints for da vinci research kit robot,'' in \emph{2023 IEEE International Conference on Robotics and Automation (ICRA)}.\hskip 1em plus 0.5em minus 0.4em\relax IEEE, 2023, pp. 776--781.

\bibitem{harris2017effect}
D.~J. Harris, S.~J. Vine, M.~R. Wilson, J.~S. McGrath, M.-E. LeBel, and G.~Buckingham, ``The effect of observing novice and expert performance on acquisition of surgical skills on a robotic platform,'' \emph{PLoS One}, vol.~12, no.~11, p. e0188233, 2017.

\bibitem{kaelbling2020foundation}
L.~P. Kaelbling, ``The foundation of efficient robot learning,'' \emph{Science}, vol. 369, no. 6506, pp. 915--916, 2020.

\bibitem{chiu2024flexible}
Z.-Y. Chiu, Y.-L. Tuan, W.~Y. Wang, and M.~Yip, ``Flexible attention-based multi-policy fusion for efficient deep reinforcement learning,'' \emph{Advances in Neural Information Processing Systems}, vol.~36, 2024.

\bibitem{zhong2019dual}
F.~Zhong, Y.~Wang, Z.~Wang, and Y.-H. Liu, ``Dual-arm robotic needle insertion with active tissue deformation for autonomous suturing,'' \emph{IEEE Robotics and Automation Letters}, vol.~4, no.~3, pp. 2669--2676, 2019.

\bibitem{liang2024real}
X.~Liang, F.~Liu, Y.~Zhang, Y.~Li, S.~Lin, and M.~Yip, ``Real-to-sim deformable object manipulation: Optimizing physics models with residual mappings for robotic surgery,'' in \emph{2024 IEEE International Conference on Robotics and Automation (ICRA)}.\hskip 1em plus 0.5em minus 0.4em\relax IEEE, 2024, pp. 15\,471--15\,477.

\bibitem{shinde2024surestep}
N.~U. Shinde, Z.-Y. Chiu, F.~Richter, J.~Lim, Y.~Zhi, S.~Herbert, and M.~C. Yip, ``Surestep: An uncertainty-aware trajectory optimization framework to enhance visual tool tracking for robust surgical automation,'' \emph{arXiv preprint arXiv:2404.00123}, 2024.

\bibitem{d2018automated}
C.~D'Ettorre, G.~Dwyer, X.~Du, F.~Chadebecq, F.~Vasconcelos, E.~De~Momi, and D.~Stoyanov, ``Automated pick-up of suturing needles for robotic surgical assistance,'' in \emph{2018 IEEE International Conference on Robotics and Automation (ICRA)}.\hskip 1em plus 0.5em minus 0.4em\relax IEEE, 2018, pp. 1370--1377.

\bibitem{ozguner2021visually}
O.~{\"O}zg{\"u}ner, T.~Shkurti, S.~Lu, W.~Newman, and M.~C. {\c{C}}avu{\c{s}}o{\u{g}}lu, ``Visually guided needle driving and pull for autonomous suturing,'' in \emph{2021 IEEE 17th international conference on automation science and engineering (CASE)}.\hskip 1em plus 0.5em minus 0.4em\relax IEEE, 2021, pp. 242--248.

\bibitem{wilcox2022learning}
A.~Wilcox, J.~Kerr, B.~Thananjeyan, J.~Ichnowski, M.~Hwang, S.~Paradis, D.~Fer, and K.~Goldberg, ``Learning to localize, grasp, and hand over unmodified surgical needles,'' in \emph{2022 International Conference on Robotics and Automation (ICRA)}.\hskip 1em plus 0.5em minus 0.4em\relax IEEE, 2022, pp. 9637--9643.

\bibitem{dharmarajan2023automating}
K.~Dharmarajan, W.~Panitch, M.~Jiang, K.~Srinivas, B.~Shi, Y.~Avigal, H.~Huang, T.~Low, D.~Fer, and K.~Goldberg, ``Automating vascular shunt insertion with the dvrk surgical robot,'' in \emph{2023 IEEE International Conference on Robotics and Automation (ICRA)}.\hskip 1em plus 0.5em minus 0.4em\relax IEEE, 2023, pp. 6781--6788.

\bibitem{shin2019autonomous}
C.~Shin, P.~W. Ferguson, S.~A. Pedram, J.~Ma, E.~P. Dutson, and J.~Rosen, ``Autonomous tissue manipulation via surgical robot using learning based model predictive control,'' in \emph{2019 International conference on robotics and automation (ICRA)}.\hskip 1em plus 0.5em minus 0.4em\relax IEEE, 2019, pp. 3875--3881.

\bibitem{kim2024surgical}
J.~W. Kim, T.~Z. Zhao, S.~Schmidgall, A.~Deguet, M.~Kobilarov, C.~Finn, and A.~Krieger, ``Surgical robot transformer (srt): Imitation learning for surgical tasks,'' \emph{arXiv preprint arXiv:2407.12998}, 2024.

\bibitem{kawaharazuka2024robotic}
K.~Kawaharazuka, K.~Okada, and M.~Inaba, ``Robotic constrained imitation learning for the peg transfer task in fundamentals of laparoscopic surgery,'' \emph{arXiv preprint arXiv:2405.03440}, 2024.

\bibitem{baek2018path}
D.~Baek, M.~Hwang, H.~Kim, and D.-S. Kwon, ``Path planning for automation of surgery robot based on probabilistic roadmap and reinforcement learning,'' in \emph{2018 15th international conference on ubiquitous robots (UR)}.\hskip 1em plus 0.5em minus 0.4em\relax IEEE, 2018, pp. 342--347.

\bibitem{nguyen2019manipulating}
N.~D. Nguyen, T.~Nguyen, S.~Nahavandi, A.~Bhatti, and G.~Guest, ``Manipulating soft tissues by deep reinforcement learning for autonomous robotic surgery,'' in \emph{2019 IEEE International Systems Conference (SysCon)}.\hskip 1em plus 0.5em minus 0.4em\relax IEEE, 2019, pp. 1--7.

\bibitem{pore2021safe}
A.~Pore, D.~Corsi, E.~Marchesini, D.~Dall’Alba, A.~Casals, A.~Farinelli, and P.~Fiorini, ``Safe reinforcement learning using formal verification for tissue retraction in autonomous robotic-assisted surgery,'' in \emph{2021 IEEE/RSJ International Conference on Intelligent Robots and Systems (IROS)}.\hskip 1em plus 0.5em minus 0.4em\relax IEEE, 2021, pp. 4025--4031.

\bibitem{scheikl2022sim}
P.~M. Scheikl, E.~Tagliabue, B.~Gyenes, M.~Wagner, D.~Dall'Alba, P.~Fiorini, and F.~Mathis-Ullrich, ``Sim-to-real transfer for visual reinforcement learning of deformable object manipulation for robot-assisted surgery,'' \emph{IEEE Robotics and Automation Letters}, vol.~8, no.~2, pp. 560--567, 2022.

\bibitem{ou2023sim}
Y.~Ou and M.~Tavakoli, ``Sim-to-real surgical robot learning and autonomous planning for internal tissue points manipulation using reinforcement learning,'' \emph{IEEE Robotics and Automation Letters}, vol.~8, no.~5, pp. 2502--2509, 2023.

\bibitem{shahkoo2023autonomous}
A.~A. Shahkoo and A.~A. Abin, ``Autonomous tissue manipulation via surgical robot using deep reinforcement learning and evolutionary algorithm,'' \emph{IEEE Transactions on Medical Robotics and Bionics}, vol.~5, no.~1, pp. 30--41, 2023.

\bibitem{karimi2024reward}
Z.~Karimi, S.-H. Ho, B.~Thach, A.~Kuntz, and D.~S. Brown, ``Reward learning from suboptimal demonstrations with applications in surgical electrocautery,'' \emph{arXiv preprint arXiv:2404.07185}, 2024.

\bibitem{varier2020collaborative}
V.~M. Varier, D.~K. Rajamani, N.~Goldfarb, F.~Tavakkolmoghaddam, A.~Munawar, and G.~S. Fischer, ``Collaborative suturing: A reinforcement learning approach to automate hand-off task in suturing for surgical robots,'' in \emph{2020 29th IEEE international conference on robot and human interactive communication (RO-MAN)}.\hskip 1em plus 0.5em minus 0.4em\relax IEEE, 2020, pp. 1380--1386.

\bibitem{su2020reinforcement}
H.~Su, Y.~Hu, Z.~Li, A.~Knoll, G.~Ferrigno, and E.~De~Momi, ``Reinforcement learning based manipulation skill transferring for robot-assisted minimally invasive surgery,'' in \emph{2020 IEEE International Conference on Robotics and Automation (ICRA)}.\hskip 1em plus 0.5em minus 0.4em\relax IEEE, 2020, pp. 2203--2208.

\bibitem{barnoy2021robotic}
Y.~Barnoy, M.~O'Brien, W.~Wang, and G.~Hager, ``Robotic surgery with lean reinforcement learning,'' \emph{arXiv preprint arXiv:2105.01006}, 2021.

\bibitem{chiu2021bimanual}
Z.-Y. Chiu, F.~Richter, E.~K. Funk, R.~K. Orosco, and M.~C. Yip, ``Bimanual regrasping for suture needles using reinforcement learning for rapid motion planning,'' in \emph{2021 IEEE International Conference on Robotics and Automation (ICRA)}.\hskip 1em plus 0.5em minus 0.4em\relax IEEE, 2021, pp. 7737--7743.

\bibitem{bendikas2023learning}
R.~Bendikas, V.~Modugno, D.~Kanoulas, F.~Vasconcelos, and D.~Stoyanov, ``Learning needle pick-and-place without expert demonstrations,'' \emph{IEEE Robotics and Automation Letters}, vol.~8, no.~6, pp. 3326--3333, 2023.

\bibitem{caianiello2023exploring}
M.~Caianiello, C.~Iacono, A.~Imperato, and F.~Ficuciello, ``Exploring the use of deep reinforcement learning algorithms for wound-approaching trajectories in robot-assisted minimally invasive surgery,'' in \emph{2023 21st International Conference on Advanced Robotics (ICAR)}.\hskip 1em plus 0.5em minus 0.4em\relax IEEE, 2023, pp. 285--290.

\bibitem{haiderbhai2024sim2real}
M.~Haiderbhai, R.~Gondokaryono, A.~Wu, and L.~A. Kahrs, ``Sim2real rope cutting with a surgical robot using vision-based reinforcement learning,'' \emph{IEEE Transactions on Automation Science and Engineering}, 2024.

\bibitem{d2022learning}
C.~D’Ettorre, S.~Zirino, N.~N. Dei, A.~Stilli, E.~De~Momi, and D.~Stoyanov, ``Learning intraoperative organ manipulation with context-based reinforcement learning,'' \emph{International Journal of Computer Assisted Radiology and Surgery}, vol.~17, no.~8, pp. 1419--1427, 2022.

\bibitem{scheikl2021cooperative}
P.~M. Scheikl, B.~Gyenes, T.~Davitashvili, R.~Younis, A.~Schulze, B.~P. M{\"u}ller-Stich, G.~Neumann, M.~Wagner, and F.~Mathis-Ullrich, ``Cooperative assistance in robotic surgery through multi-agent reinforcement learning,'' in \emph{2021 IEEE/RSJ International Conference on Intelligent Robots and Systems (IROS)}.\hskip 1em plus 0.5em minus 0.4em\relax IEEE, 2021, pp. 1859--1864.

\bibitem{fan2024learn}
K.~Fan, Z.~Chen, G.~Ferrigno, and E.~De~Momi, ``Learn from safe experience: Safe reinforcement learning for task automation of surgical robot,'' \emph{IEEE Transactions on Artificial Intelligence}, 2024.

\bibitem{huang2023guided}
T.~Huang, K.~Chen, B.~Li, Y.-H. Liu, and Q.~Dou, ``Guided reinforcement learning with efficient exploration for task automation of surgical robot,'' in \emph{2023 IEEE International Conference on Robotics and Automation (ICRA)}.\hskip 1em plus 0.5em minus 0.4em\relax IEEE, 2023, pp. 4640--4647.

\bibitem{ou2023towards}
Y.~Ou and M.~Tavakoli, ``Towards safe and efficient reinforcement learning for surgical robots using real-time human supervision and demonstration,'' in \emph{2023 International Symposium on Medical Robotics (ISMR)}.\hskip 1em plus 0.5em minus 0.4em\relax IEEE, 2023, pp. 1--7.

\bibitem{jang2016categorical}
E.~Jang, S.~Gu, and B.~Poole, ``Categorical reparameterization with gumbel-softmax,'' \emph{arXiv preprint arXiv:1611.01144}, 2016.

\bibitem{haarnoja2018soft}
T.~Haarnoja, A.~Zhou, P.~Abbeel, and S.~Levine, ``Soft actor-critic: Off-policy maximum entropy deep reinforcement learning with a stochastic actor,'' in \emph{International conference on machine learning}.\hskip 1em plus 0.5em minus 0.4em\relax PMLR, 2018, pp. 1861--1870.

\bibitem{xu2021surrol}
J.~Xu, B.~Li, B.~Lu, Y.-H. Liu, Q.~Dou, and P.-A. Heng, ``Surrol: An open-source reinforcement learning centered and dvrk compatible platform for surgical robot learning,'' in \emph{2021 IEEE/RSJ International Conference on Intelligent Robots and Systems (IROS)}.\hskip 1em plus 0.5em minus 0.4em\relax IEEE, 2021, pp. 1821--1828.

\bibitem{kazanzides2014open}
P.~Kazanzides, Z.~Chen, A.~Deguet, G.~S. Fischer, R.~H. Taylor, and S.~P. DiMaio, ``An open-source research kit for the da vinci{\textregistered} surgical system,'' in \emph{2014 IEEE international conference on robotics and automation (ICRA)}.\hskip 1em plus 0.5em minus 0.4em\relax IEEE, 2014, pp. 6434--6439.

\bibitem{richter2021robotic}
F.~Richter, J.~Lu, R.~K. Orosco, and M.~C. Yip, ``Robotic tool tracking under partially visible kinematic chain: A unified approach,'' \emph{IEEE Transactions on Robotics}, vol.~38, no.~3, pp. 1653--1670, 2021.

\bibitem{chiu2022markerless}
Z.-Y. Chiu, A.~Z. Liao, F.~Richter, B.~Johnson, and M.~C. Yip, ``Markerless suture needle 6d pose tracking with robust uncertainty estimation for autonomous minimally invasive robotic surgery,'' in \emph{2022 IEEE/RSJ International Conference on Intelligent Robots and Systems (IROS)}.\hskip 1em plus 0.5em minus 0.4em\relax IEEE, 2022, pp. 5286--5292.

\bibitem{chiu2023real}
Z.-Y. Chiu, F.~Richter, and M.~C. Yip, ``Real-time constrained 6d object-pose tracking of an in-hand suture needle for minimally invasive robotic surgery,'' in \emph{2023 IEEE International Conference on Robotics and Automation (ICRA)}.\hskip 1em plus 0.5em minus 0.4em\relax IEEE, 2023, pp. 4761--4767.

\bibitem{cheng2024putting}
H.~K. Cheng, S.~W. Oh, B.~Price, J.-Y. Lee, and A.~Schwing, ``Putting the object back into video object segmentation,'' in \emph{Proceedings of the IEEE/CVF Conference on Computer Vision and Pattern Recognition}, 2024, pp. 3151--3161.

\bibitem{kirillov2023segment}
A.~Kirillov, E.~Mintun, N.~Ravi, H.~Mao, C.~Rolland, L.~Gustafson, T.~Xiao, S.~Whitehead, A.~C. Berg, W.-Y. Lo \emph{et~al.}, ``Segment anything,'' in \emph{Proceedings of the IEEE/CVF International Conference on Computer Vision}, 2023, pp. 4015--4026.

\bibitem{garrido2014automatic}
S.~Garrido-Jurado, R.~Mu{\~n}oz-Salinas, F.~J. Madrid-Cuevas, and M.~J. Mar{\'\i}n-Jim{\'e}nez, ``Automatic generation and detection of highly reliable fiducial markers under occlusion,'' \emph{Pattern Recognition}, vol.~47, no.~6, pp. 2280--2292, 2014.

\end{thebibliography}
